\documentclass[runningheads]{llncs}

\usepackage{eccv}

\usepackage{eccvabbrv}

\usepackage{graphicx}
\usepackage{booktabs}

\usepackage[accsupp]{axessibility}  

\title{Learning Cross-hand Policies of High-DOF Reaching and Grasping}
\titlerunning{Learning Cross-hand Policies.}

\author{Qijin She\inst{1}\and
Shishun Zhang\inst{1}\and
Yunfan Ye\inst{3}\and
Ruizhen Hu\inst{2}\and
Kai Xu\inst{1}\thanks{Corresponding author. Email: kevin.kai.xu@gmail.com}
}

\authorrunning{She et al.}

\institute{National University of Defense Technology, China \and
Shenzhen University, China \and
Hunan University, China 
}

\usepackage{graphicx}
\usepackage{amsmath}
\usepackage{amssymb}
\usepackage{multirow}
\usepackage{xcolor}
\usepackage{wrapfig}
\usepackage{caption}
\usepackage{diagbox}
\usepackage{wrapfig}

\begin{document}
\maketitle
\captionsetup[table]{skip=0pt}
\captionsetup[figure]{skip=0pt}
\newcommand{\qj}[1]{{\color{black}\textbf{}#1}\normalfont}
\newcommand{\rz}[1]{{\color{black}\textbf{}#1}\normalfont}
\newcommand{\kevin}[1]{{\color{red}\textbf{}#1}\normalfont}
\newcommand{\shishun}[1]{{\color{black}\textbf{}#1}\normalfont}
\newcommand{\change}[1]{{\color{black}\textbf{}#1}\normalfont}


\begin{abstract}
    Reaching-and-grasping is a fundamental skill for robotic manipulation, 
    but existing methods usually train models on a specific gripper and cannot be reused on another gripper. 
    In this paper, we propose a novel method that can learn a unified policy model that can be easily transferred to different dexterous grippers.
    Our method consists of two stages: a gripper-agnostic policy model that predicts the displacements of pre-defined key points on the gripper, 
    and a gripper-specific adaptation model that \qj{translates these displacements into adjustments for controlling the grippers' joints.}
    \qj{
        The gripper state and interactions with objects are captured at the finger level using robust geometric representations, 
        integrated with a transformer-based network to address variations in gripper morphology and geometry.
    }
    \shishun{In the experiments, }we evaluate our method on several dexterous grippers and diverse objects, and the result shows \shishun{that} our method significantly outperforms the baseline methods.
    \shishun{Pioneering the transfer of grasp policies across dexterous grippers, our method effectively demonstrates its potential for learning generalizable and transferable manipulation skills for various robotic hands.}
    \keywords{Generlizable Dexterous Grasping \and Policy Transfer} 
\end{abstract}


\section{Introduction}
Reaching-and-grasping problem is a fundamental and crucial challenge in robotics and computer graphics. \shishun{Its core objective involves moving a gripper to approach an object and grasping it, where the prediction of the gripper's pose plays an important role.}
\shishun{In this research field, traditional methods usually separate the synthesis of grasp poses from the planning of the reaching process, leading to additional time costs in replanning and a lack of robustness. To solve this problem, learning-based methods typically train models to predict the gripper's next pose frame by frame in real time, and have achieved significant success with both two-finger grippers \cite{kalashnikov2018qt} and dexterous grippers \cite{mandikal2021learning, mandikal2022dexvip}.} However, the generalization capability of learning-based methods is limited. 
While many previous works have improved the models' generalization across objects \cite{mahler2018dex,fang2023anygrasp}, 
generalization across grippers, particularly for dexterous grippers, remains largely unexplored. 
\qj{Enabling the policy model to transfer across grippers could eliminate the need for expensive data collection and the time-consuming process of training individual models for each gripper.}

\begin{figure*}[t!]
    \centering
    \includegraphics[width=1.0\textwidth]{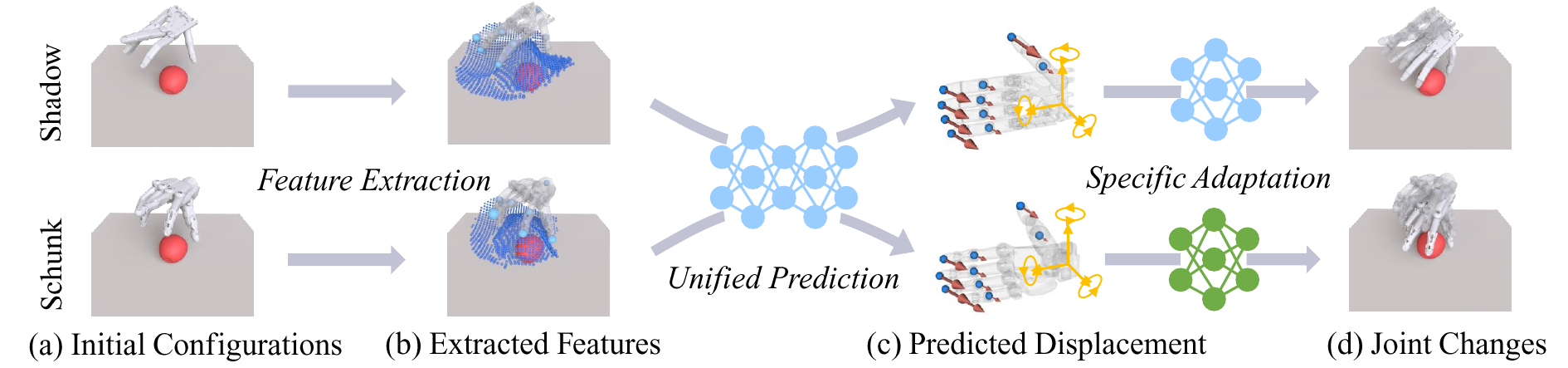}
    \setlength{\belowcaptionskip}{5pt}
    \caption {
        The overview of one step of our proposed framework.
        Given the context of the scene and the configuration of the gripper, 
        our method initially extracts gripper-agnostic features. 
        These features are uniformly sent to the policy model 
        to predict gripper-agnostic point displacements, which are forwarded to the adaptation models of various grippers for precise gripper control. 
    }
\label{fig:overview}
\end{figure*}

Prior research has aimed to create a broad grasp generalization across various grippers \cite{shao2020unigrasp, li2022efficientgrasp, li2022gendexgrasp, xu2021adagrasp}. 
However, these efforts mainly revolve around the synthesis of final grasp poses\shishun{, which implies that} 
the target pose cannot be adjusted during execution.
In this study, we aim to develop a universal High-DOF grasping policy that can be trained on a single gripper 
and subsequently transferred to other grippers with minimal effort.
\shishun{Our principal hypothesis posits that 
there are commonalities in grasping skills among grippers 
despite of their difference in geometries and morphologies. 
It is the representation
, serving as the input and the output of the policy that restricts the generalization ability of grasping skills across grippers.}
Consequently, we propose that the main challenge lies in identifying a gripper-agnostic geometric representation that can mitigate the influence of the two distinct factors:
(i)Gripper morphology: the policy model need understand the gripper's state and predict state changes, 
  which are typically represented in joint space. Due to distinct morphologies, the joint spaces of different grippers can vary significantly; 
(ii) Gripper geometry: the policy model should also receive spatial information about the gripper and the scene, which are typically represented as images or point clouds, to comprehend their spatial relationship. 
However, these representations may lead the policy model to overfit on gripper‘s geometry, potentially impacting the model's ability to generalize.

Inspired by animation systems such as IK Rig \cite{UnrealEngine2021} where animators manipulate key points on characters to alter their poses, 
we advocate the use of key points 
on and shared among dexterous grippers \shishun{(e.g. fingertip)}. 
We use \shishun{these key point positions} as the state and their displacements as the action, \shishun{to overcome} the morphological differences of the grippers. Thus, the policy \shishun{can be regarded as} ``dragging" these key points to adjust the gripper's poses. 
This pose adjustment can be accomplished through an adaptation model that converts key point displacements into gripper joint changes, simultaneously avoiding self-collisions.

\shishun{Furthermore,} to reduce the interference of gripper geometries on the policy's generalization, we introduce the Interaction Bisector Surface (IBS) \cite{zhao2014indexing} to characterize the spatial interaction between the gripper and the object. 
The IBS, computed as the Voronoi diagram between two geometric objects, has proved successful in enhancing the policy's robustness to object geometries \cite{she_sig22} in the grasping task. 
We believe that IBS should have similar robustness to gripper geometries. Through our experiments, we discovered that incorporating the IBS as an additional state representation can significantly assist policies in transferring between different grippers.
\shishun{To facilitate the policy's transfer to dexterous grippers with different fingers}, we implement the policy network based on the transformer network \cite{vaswani2017attention} for utilizing the attention mechanism to integrate information from these two representations and extract relations among fingers. 

\shishun{In the experimental section, we undertake an extensive series of performance verification and ablation studies. The findings indicate that, when compared to benchmark methods, our approach consistently produces grasps with higher quality for diverse objects. 
Moreover, our method can adapt to different grippers with minimal performance degradation, showcasing its generalization.} The main contributions of \shishun{this work can be summarized as}:
\begin{itemize}
  \item We formulate gripper-agnostic state and action representations for the grasping policy, enabling seamless transferability across different grippers;
  \item We introduce an innovative policy network designed to improve the extraction of relation among fingers and representations. 
  \item We put forward a two-stage hierarchical framework that separates the prediction of unified grasps from the control of specific grippers. 
\end{itemize}


\section{Related Work}
\label{sec:related}

\subsection{Static Generalizable Grasp Synthesis}
Grasp synthesis has been extensively analyzed from various angles. 
Existing grasp synthesis techniques can be broadly categorized into analytical and data-driven methods.
Analytical approaches employ sampling methods \cite{miller2004graspit} or optimization techniques \cite{turpin2022grasp} to search for gripper poses that ensure physical stability. 
By optimizing grasps for each object and gripper on a case-by-case basis, such methods naturally generalize across different grippers but run slowly.

Data-driven methods 
 train models that can directly predict grasp poses based on object features. 
To make these methods generalizable for different grippers, previous studies have proposed numerous cross-gripper grasp representations for prediction.
Unigrasp \cite{shao2020unigrasp} and EfficientGrasp \cite{li2022efficientgrasp} predict contact points on objects for grasping, and then employ inverse kinematics and reinforcement learning, respectively, to obtain grasps from contact points for a specific gripper.
\shishun{Following a distinct approach,} AdaGrasp \cite{xu2021adagrasp} \shishun{focuses on providing} the pre-grasp poses for grasping and then achieves stable grasps by closing the grippers. GenDexGrasp \cite{li2022gendexgrasp} generates the contact map of objects, \shishun{subsequently minimizing} the discrepancy between the actual contact map and the predicted ones, to obtain grasps for a specific gripper with a similar approach used in \cite{brahmbhatt2019contactgrasp}.
NeuralGrasp \cite{khargonkar2023neuralgrasps} \shishun{takes a unique route by learning} an implicit field defined by distances to the gripper and to the object. It generates grasps by retrieving its nearest neighbors in the latent space from a grasp database.
Nevertheless, the representations employed in the aforementioned works are specifically crafted to depict the final grasp poses and may not be well-suited for capturing the entirety of the reaching-and-grasping process.

\subsection{Kinematic Motion Retargeting}
Motion retargeting refers to the process of transferring motions from one embodied entity to another. 
There are primarily two types of motion retargeting techniques. \shishun{We summarized as learning-based and heuristic methods.}
Learning-based methods frame motion retargeting as a sequence-to-sequence generation problem \cite{villegas2018neural, aberman2019learning, aberman2020skeleton}. 
Heuristic methods employ correspondences such as matched joints \cite{penco2018robust} or key points \cite{lee1999hierarchical} between two robots to calculate the configurations of the target robot. 
Within the realm of robotic grasping, some methods have achieved real-time motion retargeting \cite{handa2020dexpilot, sivakumar2022robotic, qin2023anyteleop}, 
facilitating the transfer of motions from a human hand to a dexterous gripper. These methods are used for teleoperation and robot trajectory collection.

Indeed, while motions generated by motion retargeting methods may visually resemble the original motions, replicating them in a dynamic environment becomes challenging due to unforeseen errors. 
Hence, the kinematic motions are often employed primarily as additional data \cite{qin2022dexmv, qin2022one} or constraints \cite{mandikal2022dexvip, peng2018sfv, peng2020learning, christen2022d, zhang2024artigrasp} to train robust policies.

\subsection{Dynamic Policy Transfer}
In the pursuit of generating stable motions for new robots without the necessity for retraining, 
some studies have transferred policies directly to new robots in locomotion tasks, instead of transferring motions. \shishun{The core concept underlying} these studies is the integration of the agent's morphological information into the policy learning process. 
This morphological information can be encoded as a latent embedding upon which the policy conditions \cite{chen2018hardware, schaff2019jointly}.
\shishun{Given that} robot morphologies are often depicted as graphs, \shishun{the utilization of graph-structured neural networks \cite{scarselli2008graph} presents an effective approach to explicitly} encode the connection of different components into the policy \cite{wang2018nervenet, pathak2019learning, huang2020one}. 
\shishun{Furthermore,} transformer networks \cite{vaswani2017attention} have also provided additional options to implicitly extract relationships among various components \cite{kurin2021my, hong2021structure, gupta2022metamorph}.

However, these locomotion works are mainly designed to achieve generalization on robot morphologies without considering the impact \shishun{of} robot appearance and geometries, which cannot be ignored in contact-rich tasks such as grasping task. 
Some studies in manipulation utilize wrist-mounted cameras \cite{yang2023polybot} or image inpainting \cite{chen2024mirage} to 
mitigate the impact of the robot's appearance on policy generalization. 
Nonetheless, their approaches rely on the assumption of two-finger grippers and present difficulties in adapting to the dexterous grasping task.
GraspXL \cite{zhang2024graspxl} introduced a universal strategy for training a grasp motion policy on different dexterous grippers. However, the method still requires training separate models for each gripper and does not allow for direct policy transfer between different grippers.


\section{Methods}

Our approach is predicated on a typical grasping scenario where an object is situated on a table 
and a gripper is assigned the task of reaching and grasping it under the guidance of the policy.
Given the point clouds of the scene segmented into the foreground object and background, 
as well as the configurations of the gripper, the model is expected to output joint changes of the gripper. 
These changes are used to control the gripper's motion through PD controllers. 
The model also has to discern the task's completion when high-quality grasps form.
The environment is constructed based on PyBullet \cite{coumans2021}.

Figure \ref{fig:overview} illustrates the process by which our method controls various grippers in a single step. 
Our approach features a two-stage hierarchical model framework, distinguishing the prediction of high-level grasp motions from the specific gripper control. 
The framework consists of a unified policy model and a gripper-specific adaptation model.
The policy model, which is shared among different grippers, ingests the gripper-object interaction 
and key points. The model predicts the key point displacements, subsequently translated into gripper configuration changes through the gripper-specific adaptation model.
These three processes will be elaborated upon in Sections \ref{sec:extraction}, \ref{sec:policy} and \ref{sec:adaptation}.

\subsection{Gripper-agnostic Feature Extraction}

\label{sec:extraction}
The crux of realizing a generalized grasp policy lies in identifying a gripper-agnostic representation that encapsulates common gripper traits 
while retaining grasp-related task information. 
We combine the semantic key points and the interaction bisector surface (IBS) \cite{she_sig22}
which is the set of points equidistant to the scene and the gripper, 
to represent our policy model's state. 

Semantic key points establish a natural correspondence across different hands.
We select the fingertip point $p_k^1$ and middle phalanx point $p_k^0$ on each finger $k$, as well as the gripper's root $p_0$ on the palm. 
The middle phalanx points offer bending information of fingers, facilitating precise finger control. 
Semantic key point coordinates are calculated via forward kinematics and defined in the gripper's local coordinate system with its root to be the origin. 
We incorporate an additional 3-D vector $r$ to represent the gripper's rotation relative to the world coordinate system, defined as the initial frame of the object pose.
Thus, the full semantic key point input $s_{key}=[r, p_0, p_1^0, p_1^1..., p_k^0, p_k^1]$ comprises $6(K+1)$ dimensions where $K$
is the number of fingers.

The Interaction Bisector Surface (IBS) captures the spatial interaction between the gripper and the object. 
Computing the exact IBS is computationally intensive due to the need for Voronoi diagram extraction. 
To maintain computational efficiency, we implement the approach proposed by \cite{she_sig22} to compute sampled IBS points $\{p_{ibs}\}$ near the gripper.
We define a sphere centered at the palm with a radius $r=18cm$ and voxelize its bounding box into $v^3$ cells, where $v = 20$.
For each cell, we compute the distance from its center to both the gripper and the scene, and consider
its center as IBS points if the difference between these distances is below a threshold. 
To enhance efficiency and precision, cells distant from the potential IBS are excluded from the computation, 
and the positions of sampled IBS points are refined for closer proximity to the exact IBS surface.
An IBS illustration is available in Figure \ref{fig:overview} (b).

The sampled IBS points undergo downsampling to a fixed number $n=4096$ for network input. It's noteworthy that the origin of these points is established at the gripper's palm center.
Each IBS point $p_{b}$, with its closest points on the scene and the gripper denoted as $p_s$ and $p_g$, respectively, is associated with the following features: (i) coordinate $ c = (x,y,z) \in  R^3$; (ii) distance to the scene $d_s \in  R$; (iii) distance to the gripper $d_g \in R$;
(iv) indicator of whether $p_s$ is located on the foreground object $b_s \in \{0,1\}$; (v) one-hot indicator of the gripper component 
that $p_g$  belongs to $c_g \in  \{0, 1\}^{k+1}$;
(vi) indicator of which side of the gripper $p_g$ is located on $a_g \in [-1, 1]$. Here, $a_g = n_g \cdot d_{up}$ is the dot product of the normal direction $n_g$ of point $p_g$ on gripper in rest pose 
and the upright direction $d_{up}$ perpendicular to the palm and pointing outwards.

\subsection{Unified Policy Model}
\label{sec:policy}
The policy action is also configured to be gripper-agnostic. 
The policy model ingests IBS points $\{p_{b}\}$ and semantic key points $s_{key}$,
predicting the displacements ${\Delta p_k^i}$ of each point $p_k$, the global translation change ${\Delta p}$, the global rotation change ${\Delta r}$, 
as well as a special stop value $a_s$ used to determine task termination in cases of more than two contacts between the object and fingers.

\begin{wrapfigure}{r}{0.6\textwidth}
    \centering
	\includegraphics[width=0.5\textwidth]{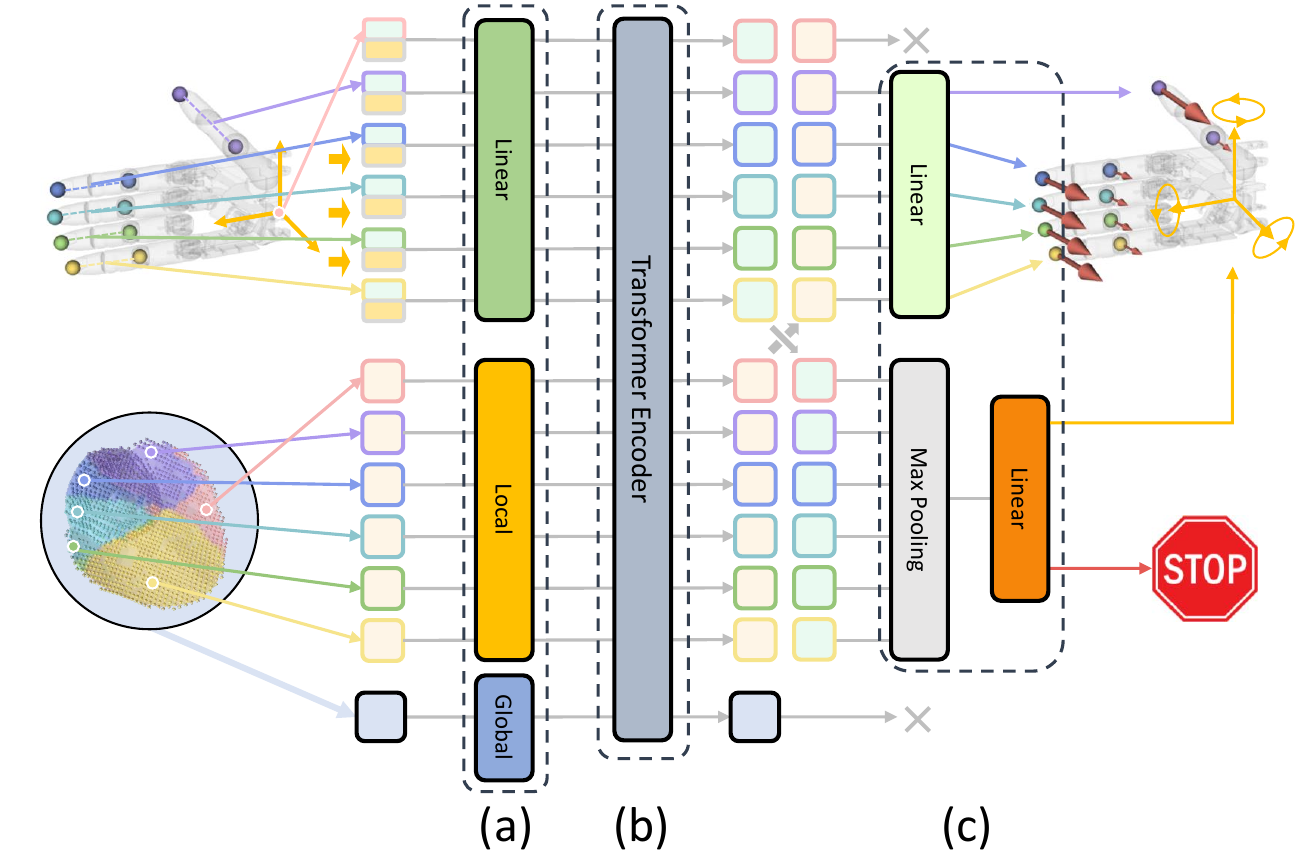}
    \caption {The policy network of our method. The network includes three components
    : (a) finger-wise feature encoder; (b) a transform encoder fusing
    information among fingers and representations; (c) finger-wise and global.
    }
    \label{fig:network}
\end{wrapfigure}

We introduce a novel transformer-based network for extracting the relationship between fingers, semantic key points, and IBS.
Figure \ref{fig:network} provides an overview of our policy model structure.
The model comprises three components: (i) Finger-wise encoders for semantic points and IBS that separately translate them into a latent feature space; 
(ii) A transformer module that executes attention over semantic key points and IBS features; (iii) Finger-wise and global feature aggregation for 
\qj{finger-wise actions and global action.}

The hand can be segmented into $K+1$ components. We categorize semantic points and IBS points according to their corresponding component 
and encode them separately using MLP and PointNet \cite{qi2017pointnet}. For each semantic key point group, we append the global rotation, resulting in a $9\-D$ vector.

Each IBS group is processed with a local PointNet network to obtain local features.
Additionally, a global PointNet computes the global features of the entire IBS.
The tokens undergo a transformer-based encoder incorporating $N$ self-attention layers. 
This attention module plays a crucial role in fostering spatial reasoning between key points and IBS defined in different coordinate systems. It effectively integrates information from other components for each hand component.
This enables individual finger action prediction, enhancing model generalizability for grippers with varying finger numbers.

The procedure involves concatenating the semantic key point embedding with the IBS embedding for each finger. Subsequently, this combined vector is projected into an output that signifies the desired semantic point displacements of the fingers within the gripper's local coordinate system.
The palm point's embedding vectors don't predict displacements but are passed into two separate MaxPooling layers along with other components' embeddings 
to predict actions impacting all fingers, i.e., the global translation and rotation change of the gripper in the world coordinate system, as well as the stop value.

\subsection{Specific Adaptation Model}
\label{sec:adaptation}
Upon determining desired key point displacements, we need to translate them into specific joint changes for gripper control, 
while avoiding finger self-collision. Though this conversion can be solved with optimization methods,
simultaneous optimization of these two objectives is time-consuming and unsuitable for real-time gripper control in our task. 
Hence, we employ a lightweight neural network, the adaptation model, to directly predict the gripper's joint changes.

The adaptation model takes the current gripper configuration, current gripper semantic points, and desired semantic key point displacements $\{\Delta p_k^i\}$ as inputs. 
It then produces the corresponding joint changes that effectuate the semantic key point displacements.
The input semantic key point positions $s_{key}'=s_{key}\setminus\{r\}$ exclude the global rotation.
The gripper configuration input $s_j=\{j\} \in R^C$ is defined as the actuated joint angles of the gripper where $C$  is the gripper's degrees of freedom. 
The model outputs the changes in gripper's joints $\{\Delta j\} \in R^C$. 
The adaptation network consists of only several MLP layers and can be efficiently trained.

\subsection{Network Training}
\label{sec:training}
The model training in our work comprises two stages: joint training and transfer training. 
Joint training aims to develop a generalizable policy model. During this stage, 
the policy model and the adaptation model are trained simultaneously, 
each with its independent loss function and no gradients are exchanged between them. 
Contrastingly, transfer training is intended to adapt the pre-trained policy model to a new gripper. 
During this stage, the policy model remains fixed while a new specific adaptation model is trained from scratch. We conduct 
$800k$ updates for both the policy and adaptation models during joint training, and 
$50k$ updates for the new adaptation model during the transfer process.

The Unified Policy Model is trained in a reinforcement learning manner.
We adopt the reward function and training strategy provided by \cite{she_sig22}.
Our reward function comprises two components: (i) a task reward for successful and stable final grasping; 
(ii) a reaching reward for preventing collision between the gripper and the scene. 
We adopt the Soft Actor-Critic \cite{haarnoja2018soft} algorithm to train our policy. 
Additional data is used to accelerate the efficiency, which is generated by the 
interpolation between static grasps and their pre-grasp poses 
and between the pre-grasp poses and the gripper's initial poses. 
We use grasps from the work \cite{liu2020deep}.

We train the Specific Adaptation Model with a self-supervised cycle loss.
Given the predicted semantic key point displacements ${\Delta p_k^i}$, current joint angles $\{j\}$, and current semantic key point positions$\{p_k^i\}$, 
the Specific Adaptation Model outputs desired joint changes $\{\Delta j\}$. 
we utilize a differentiable forward kinematic function to compute the expected semantic key point positions after joint movement.
and construct MSE Loss based on the difference between predicted key points and expected key points.  $\theta$ is the parameters of the network
\begin{equation}
    L_{point}(\theta) = \frac{1}{2} \sum_{k=1}^{K}\sum_{i=0}^{1} (e_k^i-p_k^i-\Delta p_k^i)^2,e_k^i = FK_k^i(j+\Delta j)
\end{equation}
where $FK$ is a differentiable forward kinematic function.
To avoid self-collision, i.e. the collision among different gripper components,
we incorporate a self-collision loss adapted from \cite{liu2020deep} to the original loss. 
This loss is written as:
\begin{equation}
	L(\theta) = L_{point}(\theta) + \omega L_{\text{self}}(\theta),
\end{equation}
where $\omega = 1$ is the parameter to balance different loss terms. 
The self-collision loss $L_{\text{self}}$  is defined as:
\begin{equation}
	L_{\text{self}}(\theta) = \sum_{m=1}^{L} \sum_{n=1}^{N} \max (D(p_n(j+\Delta j)), H_m(j+\Delta j)), 0),
\end{equation}
where $L$ is the number of gripper links, $N$ is the number of points $p_j(j+\Delta j)$ sampled from each link, 
$H_i(j+\Delta j)$ is the convex hull of each link, and $D$ is the signed distance from a point to a convex hull. 
More details about the self-collision loss $L_{\text{self}}$ can be found in this work \cite{liu2020deep}.


\section{Experiments}
\label{sec:result}

\subsection{Experiment Setting}
\label{sec:preparation}

\begin{table}[!t]%
	\centering
	\begin{center}
		\resizebox{\columnwidth}{!}{
			\begin{tabular}{c|c|c|c|c|c|c|c|c|c|c|c|c|c|c|c}
				\hline
				\multicolumn{1}{c|}{} & \multicolumn{5}{c|}{Features enabled} & \multicolumn{2}{c|}{Shadow(Origin)} & \multicolumn{2}{c|}{Schunk} & \multicolumn{2}{c|}{Mano} & \multicolumn{2}{c|}{Rutgers} & \multicolumn{2}{c}{Allegro}\\ \cline{2-16}
				& UNI & OCM & GCM & IBS & TR & $SR$  & $Q_1$ & $SR$ & $Q_1$ &  $SR$  & $Q_1$ &  $SR$  & $Q_1$ &  $SR$  & $Q_1$\\ \hline
				\shishun{Single} & &  & & \checkmark &                 & \textbf{72.2}$\%$ & \textbf{0.177} & - & - & - & - & - & - & - & - \\ \hline
				UNI+IBS & \checkmark & & & \checkmark&          & 68.0$\%$ & 0.168 & 54.6 $\%$ & 0.166 & 61.2$\%$ & \textbf{0.169} & 42.6$\%$ & 0.143 & - & - \\ \hline
				UNI+OCM & \checkmark & \checkmark & & &          &  50.1$\%$ & 0.152 & 41.2$\%$ & 0.164 & 45.5$\%$ & 0.166 & 38.9$\%$ & 0.145 & - & - \\ \hline
				UNI+GCM & \checkmark & & \checkmark & &          &  64.0$\%$ &  0.159 & 45.5$\%$ & 0.148 & 50.4$\%$  & 0.138 & 41.4$\%$ & 0.132 & - & -\\ \hline
				Ours & \checkmark & & & \checkmark & \checkmark& 71.3$\%$ & 0.169 & \textbf{65.3} $\%$ & \textbf{0.167} & \textbf{65.2}$\%$ & 0.167 & \textbf{54.8}$\%$ & \textbf{0.147} & \textbf{55.0}$\%$ & \textbf{0.133} \\ \hline
		\end{tabular}}
		\caption{
			Ablation Results of our method and its degraded variants using our YCB object dataset. \shishun{ ``-" means the method cannot be adapted to the gripper due to its limitations.}
		}
    \label{tab:ablation}
    \end{center}
\end{table}%

\begin{wrapfigure}{r}{0.5\textwidth}
\centering
\includegraphics[width=0.5\textwidth]{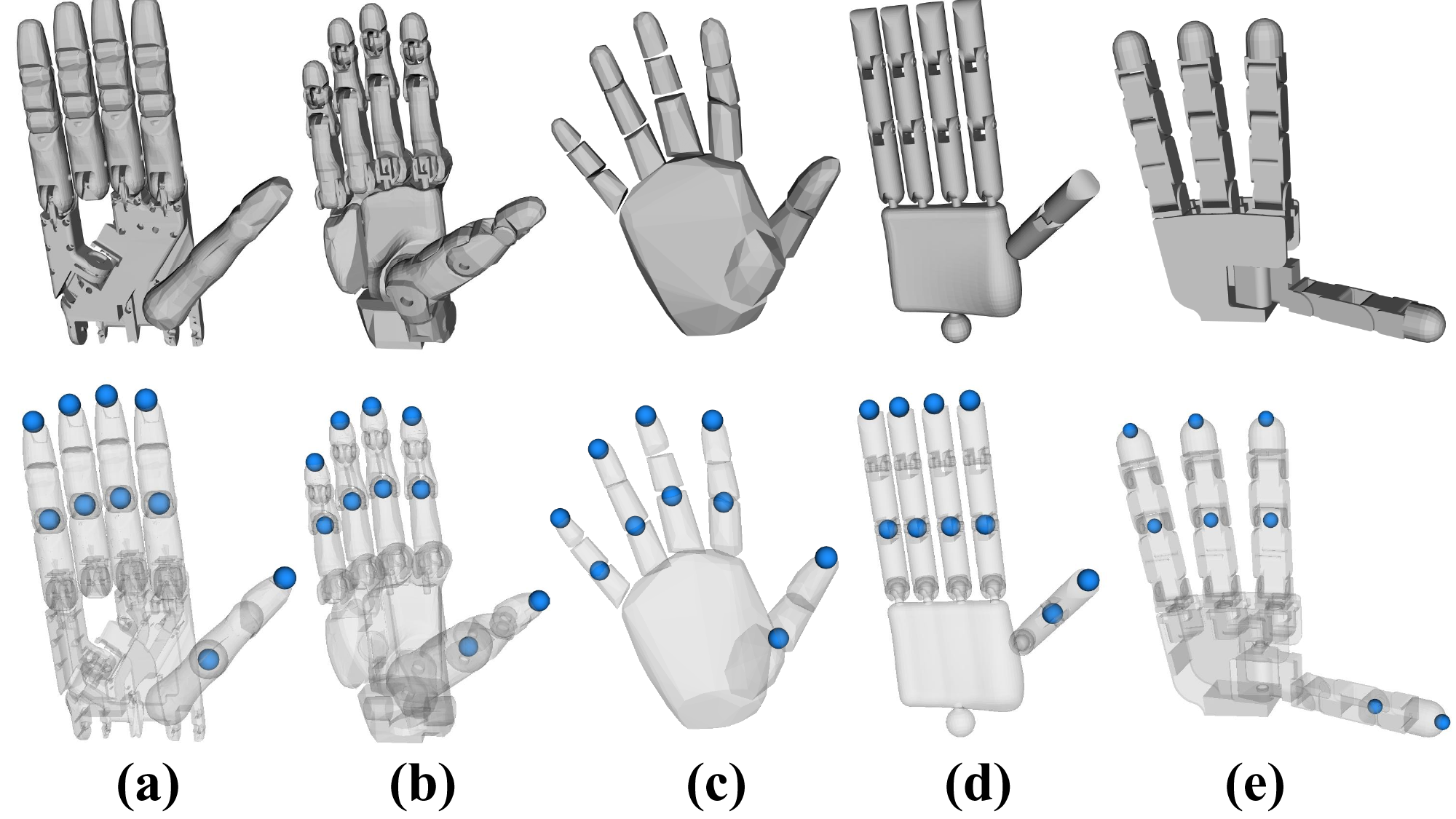}
\caption {The hands and their semantic key points. (a) Shadow; (b) Schunk; (c) Mano; (d) Rutgers; (e) Allegro.}
\label{fig:hands}
\end{wrapfigure}

We adapt several dexterous grippers for training and testing in our environment. 
These grippers include Shadow Dexterous Hand, Schunk SVH Hand, Rutgers Hand, Allegro Hand, 
and actuated virtual human hand adapted from Mano\cite{mano2017}.
Figure \ref{fig:hands} displays these hands in their rest poses. 
We select graspable objects from \cite{liu2020deep} which consists of 500 objects collected from four object datasets.
We use the objects from KIT Dataset \cite{kasper2012kit} and GD Dataset \cite{kappler2015leveraging} as training objects,
and objects from YCB Dataset \cite{calli2017yale} as test objects.
We further expand our test set with objects from ContactPose Dataset\cite{brahmbhatt2020contactpose}.
We remove the object with volume $v>1.5dm^3$ and reduce similar objects in the dataset.
In our experiment, we used Shadow Hand for policy training and other hands for adaptation tests.
To generate the initial poses of grippers in the episodes, for each object, we use its center to create a sphere with a radius $r = 20cm$. We sampled points on the upper hemisphere
as the origin of the local coordinate system of the gripper and rotated the gripper to make its palm
face the object center and its thumb point upwards. 
In the test process, we set a fixed set of initial poses for each object.

We utilize two metrics to assess policy performance. The first is the success rate of the final grasp. 
Our success criteria are adapted from the AdaGraps \cite{xu2021adagrasp}.
A grasp is deemed successful if, upon moving the gripper upwards by 
$0.6m$, the object's center of mass ascends at least $0.2m$. 
Besides, we incorporate the generalized Q1 \cite{liu2020deep} for an additional grasp stability evaluation. 
We compute its average success rate (denoted ``$\shishun{SR}$") among all objects and initial positions and the average generalized $Q_1$ values (denoted ``$Q_1$") of all successful grasps.

To validate the benefits of using IBS in representation \qj{for policy adaptation},
 we implement two types of contact map representation inspired by \cite{li2022gendexgrasp,brahmbhatt2019contactgrasp} as alternatives.
The object contact map representation (OCM) is the feature points sampled from the object's surface. 
The point features include its coordinate with the object center as its origin, its distance to the gripper,
the indicator of whether it is located in the foreground, and the one-hot indicator of the gripper component to which its closest point on the gripper belongs.
The gripper contact map representation (GCM) is the feature points sampled from the surface of the gripper. It has features similar to the ``OCM".
\qj{The point features include its coordinates with the gripper root as the origin, its distance to the scene, 
an indicator of whether its nearest point in the scene is located on the foreground object, and a one-hot indicator of the gripper component to which it belongs.}
These alternatives maintain the same point number as IBS.

\subsection{Ablation Studies on the Policy Model}
\label{sec:ablation}

We evaluate models under diverse configurations using our YCB object dataset.
Table \ref{tab:ablation} presents the results of these variants and our method.
In our ablation, we assess the impact of the following five features on the model's performance. \shishun{``Single"} means an end-to-end model with state and action defined in the gripper's joint space. ``UNI" refers to using our unified policy model and hierarchical framework.
``OCM" and ``GCM"  refer to the use of contact map representation or gripper map representation, respectively. 
``IBS" stands for the use of the IBS representation. 
``Ours" means using our transformer-based policy model instead of a naive policy model that directly concatenates features.

\paragraph{\bf Effectiveness of two-stage hierarchical framework} 
In our hierarchical framework, the unified policy model ``Uni+IBS" demonstrates its capability to control various grippers, achieving comparable performance on these hands to the end-to-end specific model \shishun{``Single"}. This is achieved despite its performance on the original hand being marginally lower.

\paragraph{\bf Importance of IBS} 
The``Uni+OCM" model, which uses feature points sampled from the object surface, underperforms in the original hand due to overfitting observed on training objects.
Despite this, it exhibits commendable transfer performance across new grippers, albeit with a slight performance drop compared to the original hand. 
On the contrary, the ``Uni+GCM" model, which employs feature points from the gripper surface as state, performs well on the original hand but struggles significantly on new hands,
\qj{This indicates that its generalizability is limited by the original grippers' geometry.}
The ``Uni+IBS" model, using an IBS representation that balances object and gripper geometry, efficiently mitigates the limitations of the preceding models.
It outperforms both ``Uni+OCM" and ``Uni+GCM" across all hands, suggesting that IBS is a viable representation for generalizable grasps encompassing various objects and grippers.

\paragraph{\bf Importance of transformer-based policy} 
By replacing the simple feature concatenation model with our transformer-based model, 
the learned policy, ``Ours", not only surpasses ``Uni+IBS"  in transfer performance 
across all hands but also successfully adapts to the four-fingered Allegro Hand. 
This indicates that our policy design not only capitalizes on the input information more effectively 
but also boasts a flexible structure that accommodates \qj{morphological} variations in the number of fingers.

\begin{table}[!t]
	\centering
    \begin{center}
    	\resizebox{\columnwidth}{!}{
        \begin{tabular}{c|c|c|c|c|c|c|c|c}
            \hline
            \multicolumn{1}{c|}{} & \multicolumn{4}{c|}{Time Cost Per Frame (ms)} & \multicolumn{2}{c|}{Grasp Performance} & \multicolumn{2}{c}{Collision}\\ \cline{2-9}
            &  Feature Extraction  & Unified Prediction & Adaptation  & Total  & $SR$ & $Q_1$ & Collision Percentage & Collision Loss \\ \hline
            OB-IK & 53.3 & 6.0 & 6.0 & 65.3 & 61.3$\%$ & 0.146 & 2.0$\%$ & 0.65\\ \hline 
            OB-IK+SC & 90.3 & 6.6 & 182.7 & 279.6 & 64.0$\%$ &  0.151 & 0.0$\%$ & 0.0 \\ \hline
            LB-IK+SC & 53.0 & 6.0 & 0.4  & 59.4 & 62.0$\%$ & 0.170 & 0.2$\%$ & 0.09 \\ \hline
        \end{tabular}}
        \caption{
        The computational efficiency and performance of our unified policy under different adaptation approaches using our YCB object dataset.
        \qj{The experiment is conducted on the Schuck Hand using a subset of initial poses.}
    }
    \label{tab:adaptation}
    \end{center}
\end{table}

\subsection{Ablation Studies on the Adaptation Module}
We perform an additional experiment that justifies our learning-based adaptation with two variants of optimization-based adaptation methods, affirming our design choice. 
The optimization method is derived from \cite{qin2022one}.
``Optimization-based Adaptation for Inverse Kinematics only (OB-IK)": 
this method employs an optimization-based approach to translate predicted key point displacements into joint changes. 
``Optimization-based Adaptation for Inverse Kinematics and self-collision avoidance (OB-IK+SC)":
the optimizer settings mirror OB-IK, with the addition of a self-collision loss to the optimization function.
``Learning-based Adaptation for IK and self-collision avoidance (LB-IK+SC)":
our approach leverages a neural network to directly predict joint changes, incorporating inverse kinematics and self-collision as loss during the training phase.

We evaluate the policy with these modules on the Schunk Hand and our YCB object set \qj{in a subset of test initial episodes} as shown in Table \ref{tab:adaptation}.
Evaluation metrics comprise the average consumption of each module in our framework, $SR$, $Q1$ as well as the percentage of collision frames 
throughout the process and the average self-collision value when such an event occurs, to illustrate the impact of self-collision avoidance loss.
The experimental results reveal that when self-collision is not accounted for, the optimization-based adaptation approach can satisfy real-time requirements, albeit with potential finger collisions. Including self-collision in the optimization function significantly escalates the algorithm's time cost, despite completely averting collision, \shishun{and this results in a running time of approximately 280ms per frame, which is difficult to meet the requirements of real-time dynamic grasping tasks.} Moreover, due to limited overall resources, the increased overhead of the optimization algorithm encroaches upon the computing resources of IBS, consequently slowing down IBS computation. 
In contrast, the learning-based method not only achieves the lowest time overhead but also prevents self-collision during the grasping process.

\subsection{Comparison to Baselines}

\begin{table}[!t]
\centering
\begin{center}
	\begin{tabular}{c|c|c|c|c|c|c}
		\hline
		\multicolumn{1}{l|}{\multirow{2}{*}{Method}} & \multicolumn{2}{c|}{Schunk} & \multicolumn{2}{c|}{Mano} & \multicolumn{2}{c}{Rutgers}\\ \cline{2-7}
		&  $SR$  & $Q_1$ & $SR$ & $Q_1$ & $SR$ & $Q_1$ \\ \hline
		MR-JM & 21.8$\%$ & 0.086 & 6.4$\%$ & 0.091 & 10.9$\%$ &	0.069 \\ \hline
		MR-KM & 35.9$\%$ & 0.136 & 33.8$\%$ & 0.124 & 33.7$\%$ & 0.126 \\ \hline
		PT-JM & 40.1$\%$ & 0.139 & 48.9 $\%$ & 0.151 & 26.9$\%$ & 0.123	\\ \hline
		PT-KM & 35.6$\%$ & 0.132 & 30.9 $\%$ & 0.132 & 31.6$\%$ & 0.141	\\ \hline
		OURS & \textbf{61.8}$\%$ & \textbf{0.167} & \textbf{63.4} $\%$ & \textbf{0.165} & \textbf{52.5}$\%$ & \textbf{0.143} \\ \hline
	\end{tabular}
\end{center}
\caption{
	Comparison results with baseline methods using both the YCB and ContactPose object datasets. 
}
\label{tab:comp}
\end{table}

\begin{figure}[!t]
    \centering
    \includegraphics[width=0.95\textwidth]{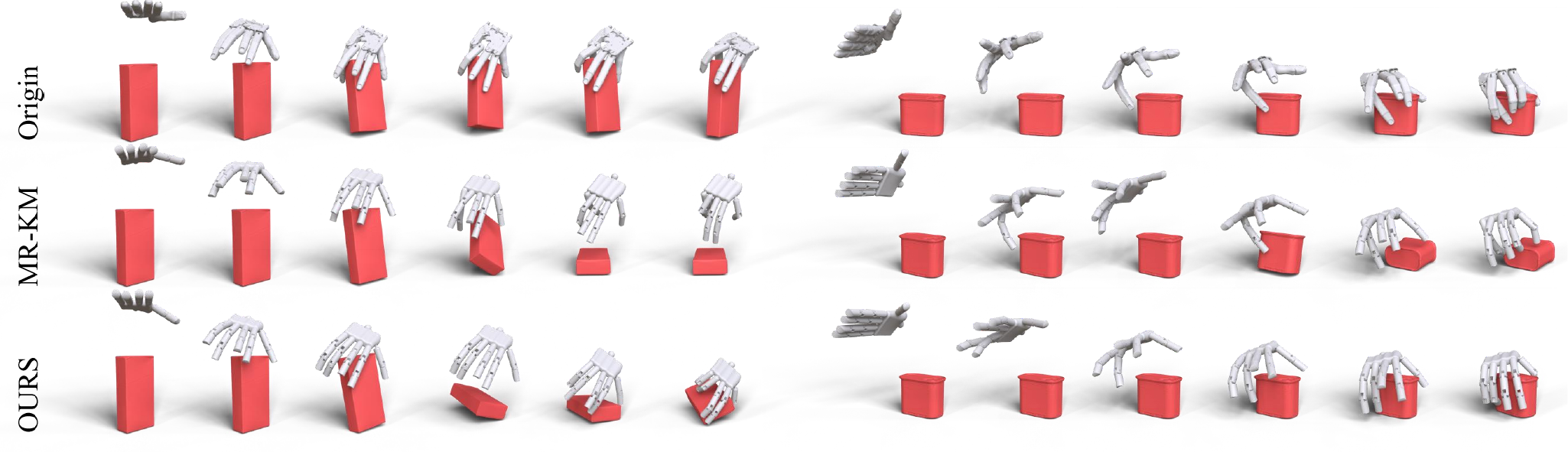}
    \caption {The visual comparisons to motion retargeting method. Our method can adapt to the pose change of the object.}
    \label{fig:mr_pt}
\end{figure}

We have implemented four baseline methods for comparison. These baselines transfer the policy trained on the specific joint space of the Shadow Hand using different strategies.
Motion Retargeting (MR) involves the offline generation of motion trajectories using a policy, which are then transferred to another hand. 
The converted trajectories are then replayed in the simulator with PD controllers.
Policy Transfer (PT) is an online process where the configuration of a new hand is mapped to the training hand, and the policy is invoked to predict the subsequent configuration for the training hand, which is then mapped back to the new hand to control it. 
This hand configuration mapping can be executed using either keypoint matching (KM) or joint matching (JM). Combining different generation and mapping methods, we have developed four baseline methods.

Table \ref{tab:comp} presents results on two object sets and three test grippers, demonstrating that our method significantly outperforms all the baseline approaches. 
Notably, ``MR-KM" surpasses ``MR-JM" in both success rate and Q1, which indicates that matching key points can aid in finding a more precise correspondence between the poses of two grippers.
The online conversion process of ``PT-JM" allows the policy to re-plan the pose, even when the conversion lacks accuracy. Consequently, it performs better than its offline version ``MR-JM", 
and even exceeds ``MR-KM" on Schunk Hand and Mano Hand. 
Figure \ref{fig:mr_pt} showcases example sequences of ``MR-KM" and ``OURS'', highlighting the advantages of replanning.
However, the expected performance improvement from ``MR-KM" to ``PT-KM" does not materialize. We attribute this primarily to the morphologies' differences of the grippers. 
The slight displacements of key points between frames on the source gripper do not trigger corresponding displacements of the key points on the target gripper during optimization.
\qj{This leads to the target gripper frequently sticking at a specific pose.}
In our method, we forecast the desired displacements of the key points on the target hand directly. This enhances the control precision over each key point, thereby facilitating the policy's adaptability across a variety of grippers.

\subsection{Verification of the Partial Observation}
To verify whether our method can work in the \qj{more realistic} partial observation setting that is closer to the real-world scenarios.
we take the incomplete point clouds captured by a depth camera as the raw input.
The camera is positioned in a fixed location under the world coordinate system, gazing upon the object from an oblique upper angle. 
We set up two test settings with different initialization strategies for gripper poses:
In ``Different gripper initialization (Diff Init)", the gripper starts from the initial pose around the object. The set of initial positions is equivalent to the test set of initial positions mentioned in Section \ref{sec:preparation}.
In ``Fixed gripper initialization (Fix Init)", the initial poses of the hand is fixed on the side of the camera, and the palm surface is parallel to the optical axis of the camera.

We test our method on these two settings as well as the original setting where the point cloud is complete. We compute the average success rate and the average generalized Q1 of all successful grasps on the YCB object set for comparison. The results are shown in Table \ref{tab:partial}.
\begin{table}[!t]
	\centering
	\begin{center}
		\begin{tabular}{c|c|c|c|c|c|c|c|c|c|c}
			\hline
			\multicolumn{1}{l|}{\multirow{2}{*}{Method}} & \multicolumn{2}{c|}{Shadow} & \multicolumn{2}{c|}{Schunk} & \multicolumn{2}{c|}{Mano} & \multicolumn{2}{c|}{Rutgers} & \multicolumn{2}{c}{Allegro}\\ \cline{2-11}
			&  $SR$  & $Q_1$ & $SR$ & $Q_1$ & $SR$ & $Q_1$ &$SR$ & $Q_1$ & $SR$ & $Q_1$\\ \hline
			Complete & \textbf{71.3$\%$} & 0.169 & \textbf{65.3$\%$} & 0.167 & \textbf{65.2$\%$} &	0.167 & 54.8$\%$ & 0.147 & \textbf{55.0$\%$} &	0.133 \\ \hline
			Diff Init & 66.8$\%$ & \textbf{0.189} & 59.2$\%$ & 0.203 & 59.9$\%$ & \textbf{0.196} & 54.8$\%$ & \textbf{0.167} & 47.1$\%$ & \textbf{0.159}\\ \hline
			Fix Init & 67.3$\%$ & 0.182 & 58.6$\%$ & \textbf{0.207} & 62.4$\%$ & 0.184 & \textbf{57.1$\%$} & 0.166 & 46.2$\%$ &	0.155	\\ \hline
		\end{tabular}
	\end{center}
	\caption{
		Results of the original setting and two partial observation test settings on the YCB object dataset. 
	}
    \label{tab:partial}
\end{table}We find our method can adapt to partial observation without much performance drop in success rates. There are two reasons why our method generalizes well to the partial observation.
First, we use IBS as a dynamic state representation. The shape of IBS remains robust even when the point cloud of the gripper and object is incomplete.
\qj{For example, when the gripper bends to grasp the object and the camera can only partially capture the object's point cloud through the gaps between fingers, the complete interaction surface can still be computed using this incomplete object point cloud.}
Secondly, our method involves learning a policy model. Even if partial observations lead to incorrect movements of the fingers, the model can adjust the gripper's behavior to rectify the error based on the new observation.
\qj{The visual example can be found in our supplimentary video.}

When comparing the performances of different dexterous hands, we observe a more significant performance drop in the Allegro Hand. This is due to its larger volume, leading to more severe occlusion between the hand and the object compared to other hands. In contrast, the Rutgers Hand, which has a relatively smaller size, experiences the smallest performance loss when provided with partial observations as input.
\qj{Additionally, we observed an interesting result. 
In cases of incomplete point clouds, the average Q1 of successful grasps using our method shows a noticeable increase.
This can be attributed to the task termination requirement in our environment, 
which mandates the finger number contacted with the objects. 
Consequently, in scenarios with incomplete point clouds, fulfilling the task termination require the contacts captured by the camera.
It is possible that other fingers make contact with parts of the object not visible in the captured point cloud. 
Since the generalized Q1 value is closely linked to the number of contacts between the fingers and the object, grasps under partial observations result in a higher Q1 value.
}

\subsection{Visualization of Grasping Processes}

\begin{figure}[!t]
    \centering
    \includegraphics[width=0.95\textwidth]{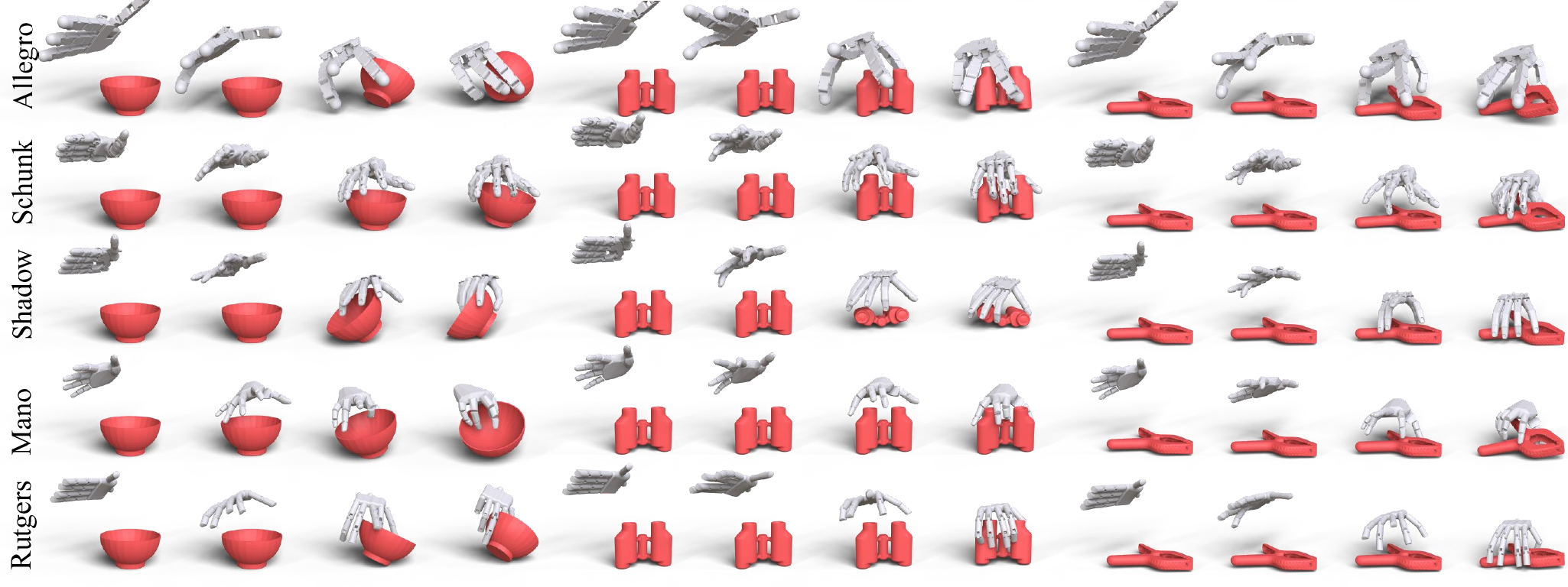}
    \caption {The visual results of our method on different grippers (up to down) and objects (left to right). For each case, we show the
    initial configurations of grippers and two sampled frames during the
    reaching process with the final grasping.}
    \label{fig:gallery}
\end{figure}

Figure \ref{fig:gallery} illustrates the visual results of our method, 
showcasing the control of grippers to grasp various objects with smooth motions.
\qj{The complete reaching processes and additional examples can be found in our supplementary video.}


\section{Conclusion}
To the best of our knowledge, our method represents the first attempt to transfer grasp policies across various dexterous grippers. 
Our approach adopts a two-stage hierarchical model framework, which separates the prediction of grasp key points' motions from the control of specific grippers. 
This, along with the use of a unified gripper-agnostic state and action designs, as well as a novel policy network, enables our method to be easily adaptable to various dexterous grippers while achieving exceptional performance.

However, our method still has some limitations. First, our method cannot transfer the policy trained on an anthropomorphic hand (e.g. Shadow Hand) to non-anthropomorphic hands (e.g. Robotiq 3F) because of large differences in topology.
We believe learning a more flexible correspondence across grippers would make our method more generalizable, and would like to leave this for future exploration. 
\qj{Moreover, 
in our experiments, we standardized the physical parameters for various dexterous hands, such as mass and joint friction. Adapting the algorithm to account for diverse physical parameters across different hands is essential for the successful implementation of our method in real-world settings.}

\bibliography{bibliography.bib}  
\bibliographystyle{splncs04}
\end{document}